\documentclass[letterpaper]{article} 
\usepackage{aaai23}  
\usepackage{times}  
\usepackage{helvet}  
\usepackage{courier}  
\usepackage[hyphens]{url}  
\usepackage{graphicx} 
\usepackage{natbib}  
\usepackage{caption} 
\usepackage{amsmath,amssymb,amsfonts}
\usepackage{multirow}
\usepackage{booktabs}
\usepackage{algorithm,algpseudocode}
\usepackage{newfloat}
\usepackage{listings}
\DeclareCaptionStyle{ruled}{labelfont=normalfont,labelsep=colon,strut=off} 
\floatstyle{ruled}
\newfloat{listing}{tb}{lst}{}
\floatname{listing}{Listing}
\usepackage{color}

\title{Delving into the Adversarial Robustness of Federated Learning}

\author{
Jie Zhang$^{1*}$\quad Bo Li$^{2*\ddagger}$ \quad Chen Chen$^{3}$ \quad Lingjuan Lyu$^{3\ddagger}$\quad \\ \textbf{Shuang Wu}$^{2}$\quad  \textbf{Shouhong Ding}$^{2}$\quad \textbf{Chao Wu}$^{1\ddagger}$
}
\affiliations{
    \textsuperscript{\rm 1}Zhejiang University \qquad 
    \textsuperscript{\rm 2}Youtu Lab, Tencent \qquad
    \textsuperscript{\rm 3}Sony AI \qquad
    \\ 
    {$\{$zj$\_$zhangjie, chao.wu$\}$@zju.edu.cn} \\ 
{$\{$libraboli, calvinwu, ericshding$\}$@tencent.com, {$\{$chen.chen, Lingjuan.Lv$\}$}@sony.com
}}

\usepackage{bibentry}

\begin{document}

\maketitle
\renewcommand{\thefootnote}{\fnsymbol{footnote}}
\footnotetext[1]{Equal contribution. Work done during Jie Zhang's internship at Tencent Youtu Lab and partly done at Sony AI.}
\footnotetext[3]{Corresponding author.}

\begin{abstract}
  In Federated Learning (FL), models are as fragile as centrally trained models against adversarial examples. However, the adversarial robustness of federated learning remains largely unexplored. This paper casts light on the challenge of adversarial robustness of federated learning. To facilitate a better understanding of the adversarial vulnerability of the existing FL methods, we conduct comprehensive robustness evaluations on various attacks and adversarial training methods. Moreover, we reveal the negative impacts induced by directly adopting adversarial training in FL, which seriously hurts the test accuracy, especially in non-IID settings. 
  In this work, we propose a novel algorithm called Decision Boundary based Federated Adversarial Training (DBFAT), which consists of two components (local re-weighting and global regularization) to improve both \textbf{accuracy} and \textbf{robustness} of FL systems. 
  Extensive experiments on multiple datasets demonstrate that DBFAT consistently outperforms other baselines under both IID and non-IID settings. 
\end{abstract}

\section{Introduction}
Nowadays, end devices are generating massive amounts of potentially sensitive user data, raising practical concerns over security and privacy. Federated Learning (FL)~\cite{mcmahan2017communication} emerges as a 
privacy-aware learning paradigm that allows multiple clients to collaboratively train neural networks 
without 
revealing their raw data. Recently, FL has 
attracted increasing attention from different areas, including medical image analysis~\cite{DBLP:conf/cvpr/Liu00DH21,DBLP:conf/miccai/ChenZYY21}, recommender systems~\cite{DBLP:conf/aaai/LiangP021,DBLP:conf/sigir/LiuXYFZM21}, natural language processing~\cite{DBLP:conf/emnlp/ZhuWHX20,DBLP:conf/emnlp/WangDMWLLHMRD21}, etc.

\begin{table*}[t]
\centering
\caption{The accuracy (\%) is tested under PGD-40 attack~\cite{madry2017towards}. For MNIST, FMNIST, CIFAR10, ImageNet-12, CIFAR100, and Tiny-ImageNet, the perturbation bound is $\{0.3,32/255,0.031,0.031,0.031,0.031\}$
, respectively. $A_{cln}$ and $A_{rob}$ refer to clean accuracy and robust accuracy.}
\label{fra}
\scalebox{0.80}{
\begin{tabular}{c|c|c|c|c|c|c|c}
\toprule
Type & Dataset & MNIST & FMNIST & ImageNet-12 & CIFAR10 & CIFAR100 & Tiny-ImageNet \\
\midrule
\multirow{2}{*}{Centralized} & $A_{cln}$ & 99.42 & 92.47 & 78.96 & 94.26 & 86.93 & 57.93 \\
 & $A_{rob}$ & 0 & 0 & 0 & 0 & 0 & 0 \\
 \midrule
\multirow{2}{*}{Federated} & $A_{cln}$ & 99.01 & 88.51 & 71.65 & 85.81 & 81.28 & 49.79 \\
 & $A_{rob}$ & 0 & 0 & 0 & 0 & 0 & 0 \\
 \bottomrule
\end{tabular}}
\end{table*}

Prior studies have demonstrated that neural networks are vulnerable to evasion attacks by adversarial examples~\cite{goodfellow2014explaining} during inference time. The goal of inference-time adversarial attack~\cite{DBLP:conf/mm/LiXWDLH21,Chen_2022_CVPR,Zhang_2022_CVPR,DBLP:conf/eccv/ChenLWXDZ22} is to 
damage the global model by adding a carefully generated imperceptible perturbation on the test examples.
As shown in Table~\ref{fra}, federated models are as fragile to adversarial examples as centrally trained models (i.e.\, zero accuracy under PGD-40 attack~\cite{madry2017towards}). Hence, it is also important to consider how to defend against adversarial attacks in federated learning.


There are several works that aim to deal with adversarial attacks in FL~\cite{DBLP:conf/icml/ZhangLLX0DW22,zhang2022dense}, i.e, federated adversarial training (FAT)~\cite{zizzo2020fat,hong2021federated,shah2021adversarial,chen2022gear,chencalfat2022}. \cite{zizzo2020fat} and \cite{hong2021federated} proposed to conduct adversarial training (AT) on a proportion of clients but conduct plain training on other clients. \cite{shah2021adversarial} investigated the impact of local training rounds in FAT. 
Nevertheless, these methods all ignore the issue that the clean accuracy of federated adversarial training is very low.

\begin{figure*}[t]
  \centering
  \includegraphics[width=15cm]{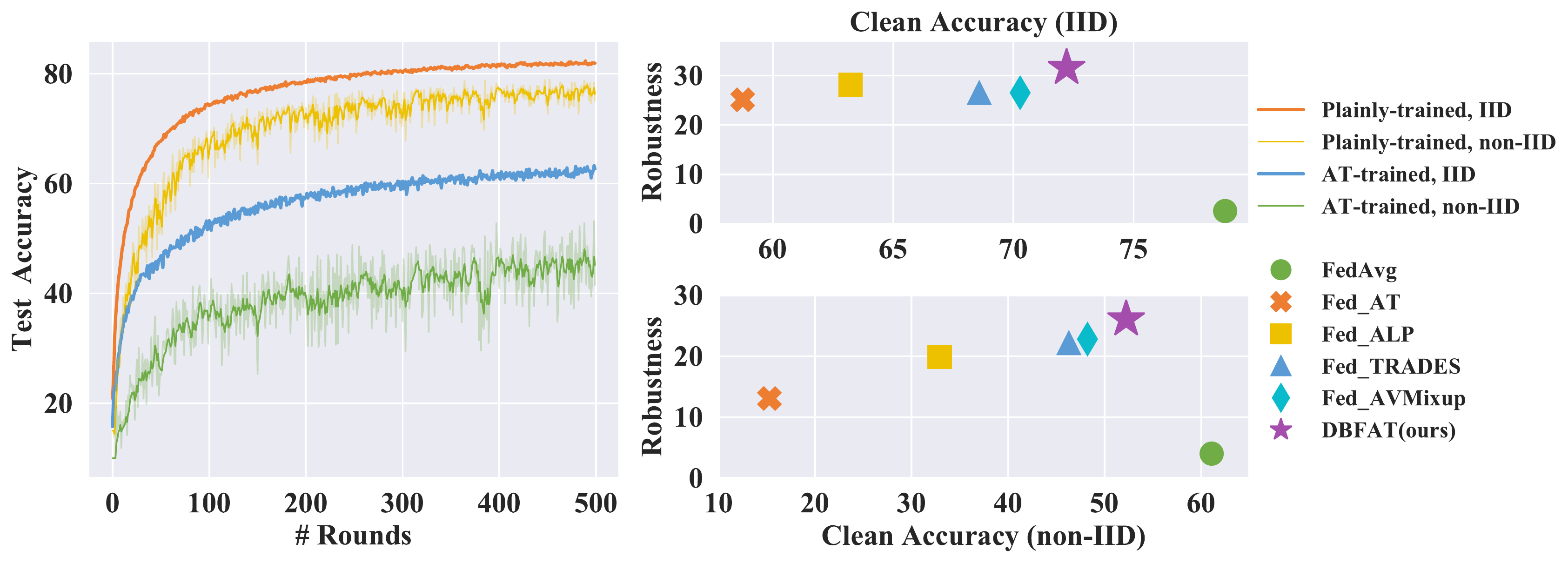} 
  \caption{\textbf{Left:} Test accuracy reduces for plainly trained model and adversarially trained model under non-IID data. Meanwhile, adversarial training hurts the 
  performance. 
  \textbf{Right:} Evaluations on CIFAR10 for both accuracy and robustness, including several state-of-the-art defense methods combined with FL. Our method outperforms existing baselines on both metric dimensions. 
  }
  \vspace*{-0.5cm}
  \label{fig1}
\end{figure*}

  



To further show the problems 
of federated adversarial training, 
we first begin with 
the comparison between the plainly-trained models and AT-trained~\cite{madry2017towards} models in both the IID (Independent and Identically Distributed) and non-IID FL settings, measured by clean accuracy \textbf{$A_{cln}$} and robust accuracy \textbf{$A_{rob}$}, respectively. 
We show the test accuracy of plain training and adversarial training (AT) on CIFAR10 dataset under both IID and non-IID FL settings in Fig.~\ref{fig1} (left sub-figure). 
We summarize some valuable observations 
as follows: 
1) Compared with the plainly-trained models, AT-trained models achieve a lower accuracy, which indicates that directly adopting adversarial training in FL can hurt $A_{cln}$; 
2) $A_{cln}$ drops heavily for both the plainly-trained models and AT-trained models under non-IID distribution, which is exactly the challenge that typical federated learning with heterogeneous data encountered~\cite{zhao2018federated};
3) The performance of AT-trained models with non-IID data distribution decrease significantly compared with IID data distribution. 
Motivated by these observations, we focus on 
improving both adversarial robustness and clean accuracy of adversarial training in FL, i.e., we aim to increase $A_{cln}$ while keeping 
$A_{rob}$ as high as possible.



To achieve this goal, in this paper, we investigate the impact of decision boundary, which can greatly influence the performance of the model in FAT. Specifically, 1) we apply adversarial training with a re-weighting strategy in local update to get a better $A_{rob}$. Our method takes the limited data of each client into account,  those samples that are close to/far from the decision boundary are assigned larger/smaller weight. 2) Moreover, since the global model in FL has a more accurate decision boundary through model aggregation, we take advantage of the logits from the global model and introduce a new regularization term to increase $A_{cln}$. This regularization term aims to alleviate the accuracy reduction across distributed clients.




We conclude our major contributions as follows:
\begin{itemize}
    \item We conduct systematic studies on the adversarial robustness of FL, and provide valuable observations from extensive experiments. 
\item We reveal the negative impacts of adopting adversarial training in FL, and then propose an effective algorithm called Decision Boundary based Federated Adversarial Training (DBFAT), which utilized local re-weighting and global regularization to improve both the accuracy and robustness of FL systems.

\item Extensive experiments on multiple datasets demonstrate that our proposed DBFAT consistently outperforms other baselines under both IID and non-IID settings. We present the performance of our method in Fig.~\ref{fig1} (right sub-figure), which indicates the improvement in both robustness and accuracy of adversarial training in FL.
\end{itemize}


\begin{table*}[t]
  \vspace*{-0.1cm}
  \caption{An empirical study on the adversarial robustness of FL, measured by various combination of defense methods and FL algorithms. We report the clean accuracy and robust accuracy, respectively. Best results are in bold.}
  \centering
  \label{tb2}
  \scalebox{0.75}{ 
      \begin{tabular}{c|cccccccc|cccccccc}
          \toprule
          Type & \multicolumn{7}{c}{IID} &  & \multicolumn{7}{c}{Non-IID} &  \\
          \midrule
          Methods & \multicolumn{2}{c}{FedAvg} & \multicolumn{2}{c}{FedProx} & \multicolumn{2}{c}{FedNova} & \multicolumn{2}{c|}{Scaffold} & \multicolumn{2}{c}{FedAvg} & \multicolumn{2}{c}{FedProx} & \multicolumn{2}{c}{FedNova} & \multicolumn{2}{c}{Scaffold} \\
          \midrule
          Performance & $A_{cln}$ & $A_{rob}$ & $A_{cln}$ & $A_{rob}$ & $A_{cln}$ & $A_{rob}$ & $A_{cln}$ & $A_{rob}$ & $A_{cln}$ & $A_{rob}$ & $A_{cln}$ & $A_{rob}$ & $A_{cln}$ & $A_{rob}$ & $A_{cln}$ & $A_{rob}$ \\
          \midrule
          PGD-AT & 57.99 & 31.95 & 58.17 & 32.06 & 58.45 & 31.74 & 56.84 & 29.26 & 46.84 & 26.79 & 48.03 & 27.46 & 46.95 & 26.54 & 42.44 & 27.19 \\
          ALP & 62.81 & 31.84 & 62.88 & 31.20 & 62.91 & 31.79 & 60.30 & 29.58 & 56.16 & \textbf{28.78} & 55.79 & \textbf{29.06} & 55.80 & \textbf{29.18} & 48.29 & 26.56 \\
          TRADES & 64.94 & \textbf{32.93} & 64.29 & 32.97 & 64.46 & 33.29 & 63.14 & \textbf{33.58} & 60.94 & 27.06 & 61.05 & 27.94 & 60.34 & 28.78 & 59.53 & 27.78 \\
          MMA & 65.14 & 30.29 & 63.65 & 31.29 & \textbf{65.27} & 29.31 & 64.28 & 32.98 & 59.69 & 28.64 & 60.17 & 28.09 & 61.03 & 28.47 & 61.53 & 28.13 \\
          AVMixup & \textbf{66.14} & 32.27 & \textbf{65.12} & \textbf{33.19} & 65.14 & \textbf{33.75} &\textbf{65.11} & 33.24 & \textbf{61.17} & 28.56 & \textbf{61.47} & 28.34 & \textbf{62.04} & 28.12 & \textbf{61.91} & \textbf{28.81} \\
          \bottomrule
          \end{tabular}
  }
\end{table*}


\section{Related Works}
\subsubsection{Federated Learning.} 
Following the success of DNNs in various tasks~\cite{DBLP:conf/ijcai/0061STSS19,DBLP:conf/aaai/LiSG19,,huang2022fastdiff,huang2022singgan,9616392_Dong}, FL has attracted increasing attention.
A recent survey has pointed out that existing FL systems are vulnerable to various attacks that aim to either compromise data privacy or system robustness~\cite{lyu2022privacy}. In particular, robustness attacks can be broadly classified into training-time attacks (data poisoning and model poisoning) and inference-time attacks (evasion attacks, 
i.e., using adversarial examples to attack the global model during inference phase). In FL, the architectural design, distributed nature, and data constraints can bring new threats and failures
~\cite{kairouz2021advances}. 



  \subsubsection{Adversarial Attacks.}
  The white-box attacks have access to the whole details of threat models, including parameters and architectures. Goodfellow et al. \cite{goodfellow2014explaining} introduced the Fast Gradient Sign Method (FGSM) to generate adversarial examples, which uses a single-step first-order approximation to perform gradient ascent. Kurakin et al. \cite{kurakin2017adversarial} iteratively applied FGSM with a small step-size to develop a significantly stronger multi-step variant, called Iterative FGSM (I-FGSM). Based on these findings, more powerful attacks have been proposed in recent years including MIM~\cite{dong2018boosting}, PGD~\cite{madry2017towards}, CW~\cite{carlini2017towards}, and AA~\cite{croce2020reliable}. 
\subsubsection{Adversarial Training.}
Adversarial training has been one of the most effective defense strategies against adversarial attacks. Madry et al. \cite{madry2017towards} regarded adversarial training as a min-max formulation using empirical risk minimization under PGD attack. Kannan et al. \cite{kannan2018adversarial} presented adversarial logit pairing (ALP), a method that encourages logits for pairs of examples to be similar, to improve robust accuracy. To quantify the trade-off between accuracy and robustness, Zhang et al. \cite{zhang2019theoretically} introduced a TRADES loss to achieve a tight upper bound on the gap between clean and robust error. Based on the margin theory and soft-labeled data augmentation, Ding et al. \cite{ding2020mma} proposed Max-Margin Adversarial (MMA) training and Lee et al.~\cite{lee2020adversarial} introduced Adversarial Vertex mixup (AVmixup).

\subsubsection{Federated Adversarial Training.}
In terms of the adversarial robustness, Zizzo et al.~\cite{zizzo2020fat} investigated the
effectiveness of the federated adversarial training protocol for idealized federated settings, and showed the performance of their models in a traditional centralized setting and a distributed FL scenario. 
Zhou et al.~\cite{zhou2022adversarial} decomposed the aggregation error of the central server into bias and variance. 
However, all these methods sacrificed clean accuracy (compared to plainly trained models) to gain robustness.
In addition, certified defense~\cite{chen2021certifiablyrobust} against adversarial examples in FL is another interesting direction, which will be discussed in the future.

\section{Adversarial Robustness of FL}
In this section, we briefly define the goal of federated 
adversarial training. Then we conduct a systematic study on some popular federated learning algorithms with the combination of various adversarial training methods and evaluate their robustness under several attacks.  
Besides, we further reveal the challenges of adversarial training in non-IID FL. 
\subsection{Problem Definition}

In typical federated learning, 
training data are distributed across all the $K$ clients, and there is a central server managing model aggregations and communications with clients. In general, federated learning attempts to minimize the following optimization:
\begin{equation} 
    \min _{w} f(w)=\sum_{k=1}^{K} \frac{n_k}{n} F_{k}(w). 
\end{equation}


Here, we denote that the global approximate optimal is a sum of local objectives weighted by the local data size $n_k$, and $n$ is the total data size of all clients that participate in a communication round. Moreover, each local objective measures the empirical risk over possibly different data distributions $D_k$, which can be expressed as: 
\begin{equation}
    F_{k}(w):=\mathbb{E}_{x_{k} \sim \mathcal{D}_{k}}\left[f_{k}\left(w ; x_{k}\right)\right].
\end{equation}

Let $x$ denote the original image, $x^{adv}$ denote the corresponding adversarial example, and $\delta$ denote the perturbation added on the original image, then $x^{adv}=x+\delta$. To generate powerful adversarial examples, we attempt to maximize the loss $L(x+\delta;w)$, where $L$ is the loss function for local update.

To improve the robustness of the neural networks, many adversarial defense methods have been proposed. 
Among them, adversarial training~\cite{carlini2017towards} is one of the most prevailing and effective algorithms. Combined with adversarial training, the local objective 
becomes solving the following min-max optimization problem:
\begin{equation}
    \label{eq1}
       F_{k}(w)=\min \mathbb{E}_{x_{k} \sim \mathcal{D}_{k}} \left[ \max _{\|x^{adv}-x\|_{\infty} \leq \delta} L(w, x^{adv}, y)\right].
 \end{equation}

 The inner maximization problem aims to find effective adversarial examples that achieve a high loss, while the outer optimization updates local models to minimize training loss. 
 
 In this work, we conduct a systematic study on several state-of-the-art FL algorithms including FedAvg~\cite{mcmahan2017communication}, FedProx~\cite{li2018federated}, FedNova~\cite{wang2020tackling} and Scaffold~\cite{karimireddy2020scaffold}, and explore their combinations with AT methods to defend against adversarial attacks. We report detailed results in Table~\ref{tb2}, here robustness is averaged over four popular attacks (FGSM~\cite{kurakin2017adversarial}, MIM~\cite{dong2018boosting}, PGD~\cite{madry2017towards}, and CW~\cite{carlini2017towards}). Besides, we implement some 
 prevailing adversarial training methods including PGD\_AT~\cite{madry2017towards}
 , TRADES~\cite{zhang2019theoretically}, ALP~\cite{kannan2018adversarial}, MMA~\cite{ding2020mma} and AVMixup~\cite{lee2020adversarial}. 
 We observe that there is no federated adversarial learning algorithm that can outperform all the others in all cases. Moreover, the clean accuracy drops heavily under non-IID distribution. 
 As such, we are motivated to develop a more effective method. Due to the similar performance of these FL methods observed from Table~\ref{tb2}, we design our method based on FedAvg -- a representative algorithm in FL.
 

\subsection{Adversarial Traning with non-IID Data}


Federated learning faces the statistical challenge in real-world scenarios. The IID data makes 
the stochastic gradient as an unbiased estimate of the full gradient~\cite{mcmahan2017communication}. However, the clients are typically highly heterogeneous with various kinds of non-IID settings, such as label 
skewness and feature 
skewness~\cite{li2021federated}. According to previous studies~\cite{wang2020tackling,karimireddy2020scaffold}, the non-IID data settings can degrade the effectiveness of the deployed model. 

\begin{figure}
  \centering
  \includegraphics[width=.4\textwidth]{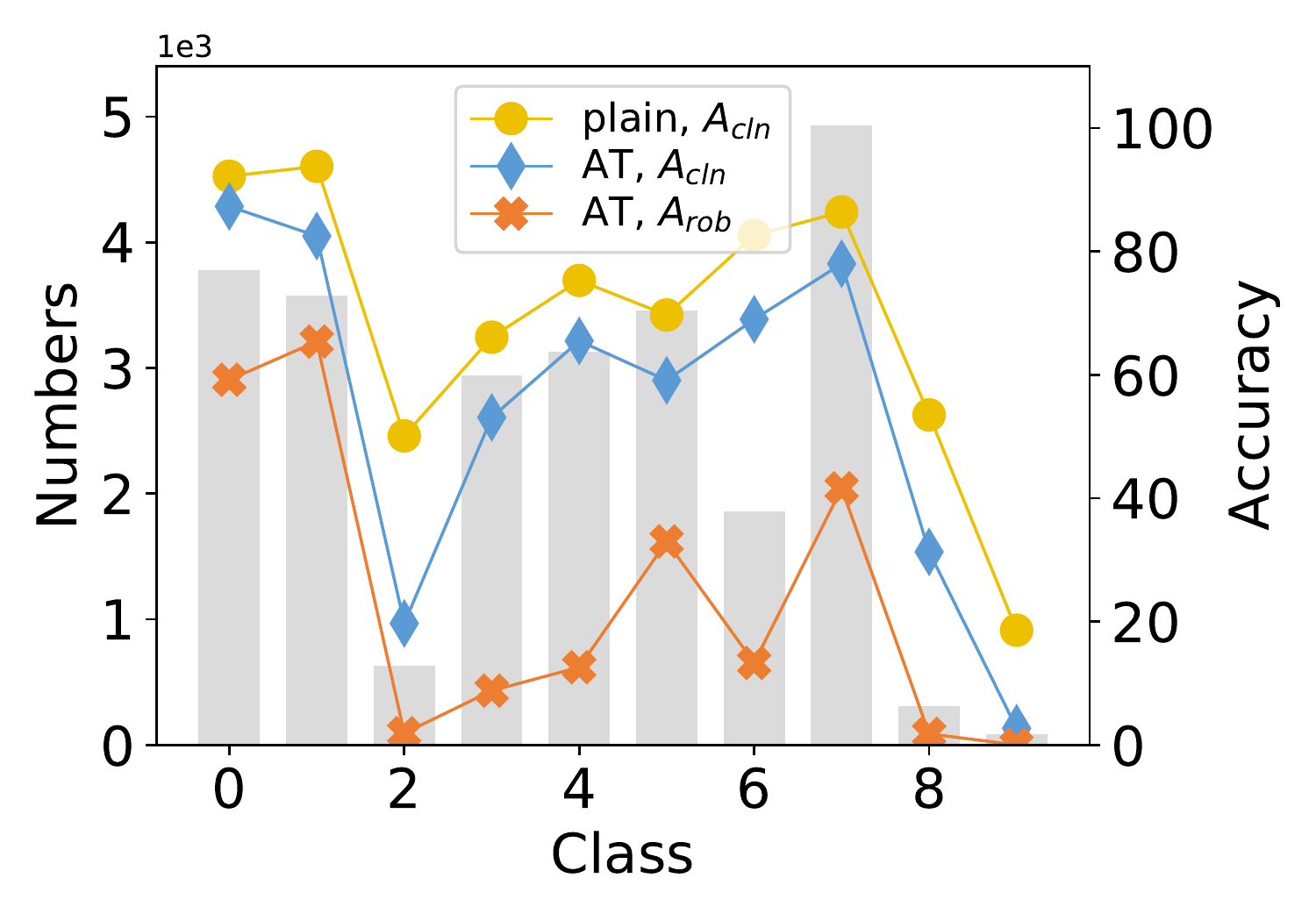}
  \caption{Test accuracy on a randomly selected client.}
  \label{fig:acc_rob}
\end{figure} 
Similarly, due to the non-IID data, the performance of AT may vary widely across clients. 
To better understand the challenge of adversarial training with non-IID data, we examine the performance of both clean accuracy and robustness on a randomly selected client and report the results in Fig.~\ref{fig:acc_rob}. Observed from Fig.~\ref{fig:acc_rob}, we can find that: 1) $A_{cln}$ on the plainly trained model drops from majority classes to minority classes, which is exactly what traditional imbalanced learning attempts to solve; 2) A similar decreasing tendency reasonably occurs in $A_{rob}$. It is obvious that adopting adversarial training in federated learning with non-IID data is more challenging.

\begin{figure}[t]
    \centering
    \includegraphics[scale=0.24]{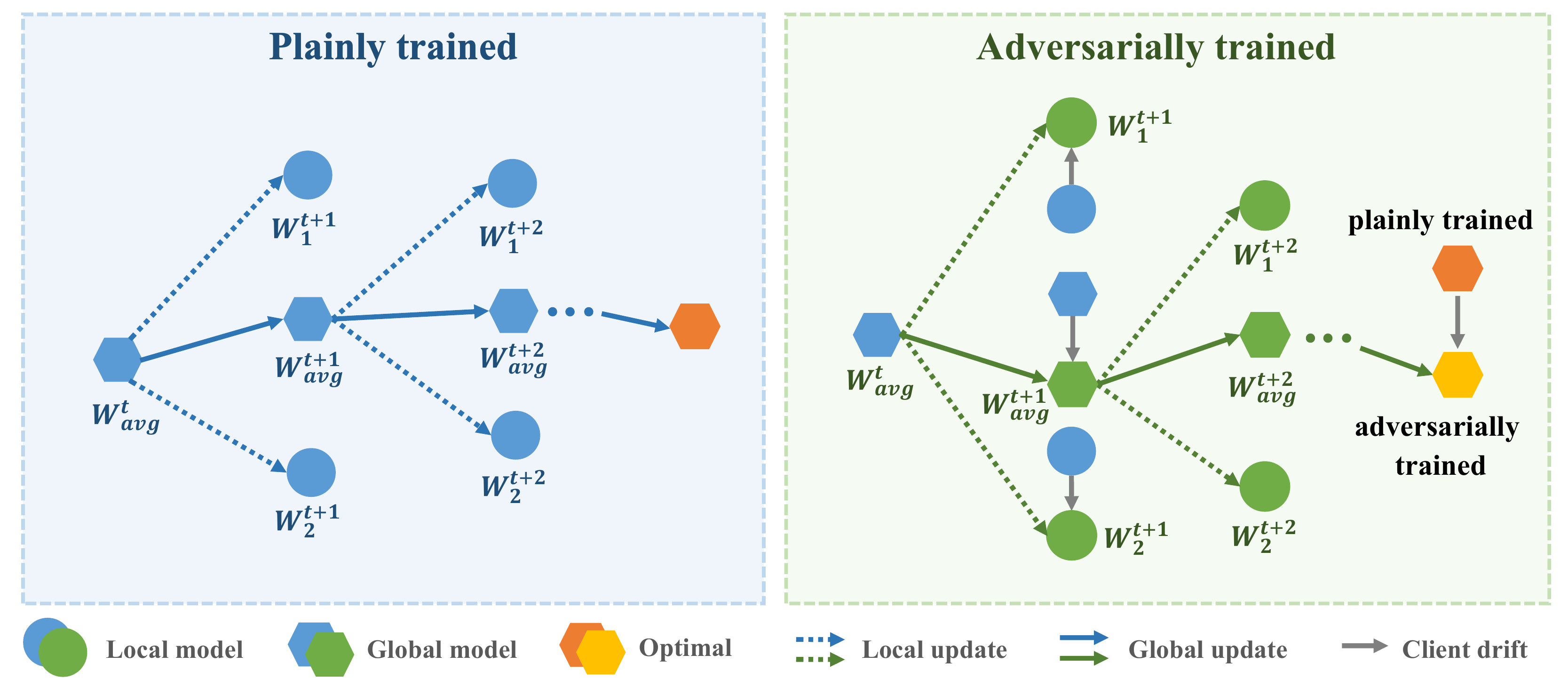}
    \caption{ 
      Plain training and adversarial training under non-IID setting. Compared with plainly trained situation, the aggregation of adversarially trained models can lead to a more biased model which enlarges  accuracy gap. Consequently, it results in poor consistency between different clients.
      }
    \label{fig:pla_at}
    \vspace{-0.4cm}
\end{figure}

According to 
above observations, we 
conjecture that AT-trained local models with imbalanced data lead to a more biased decision boundary than plainly trained ones. Since adversarial examples need a larger number of epochs to achieve near-zero error~\cite{zhang2021geometryaware}, it becomes harder to fit adversarial examples than clean data. However, for the local client itself, imbalanced clean data generates imbalanced adversarial examples, making it more difficult for training and enlarging the accuracy gap, which can reduce the performance both in accuracy and robustness. In Fig.~\ref{fig:pla_at}, we also show the differences between plain training and adversarial training in federated settings. Compared with the plainly trained models, the aggregation of adversarially trained models can enlarge the accuracy gap, which results in poor consistency between different clients. To overcome this problem, we propose a novel method to utilize local re-weighting and global regularization to improve both the accuracy and robustness of FL systems.

\section{Methodology}
\label{sec4}
  

The generalization performance of a neural network is closely related to its decision boundary. However, models trained in the federated setting are biased compared with the centrally trained models. This is mainly caused by heterogeneous data and objective inconsistency between clients~\cite{kairouz2021advances}. Moreover, a highly skewed data distribution can lead to an extremely biased boundary~\cite{wang2020tackling}. We tackle this problem in two ways: 1) locally, we take full advantage of the limited data on the distributed client; 2) globally, we utilize the information obtained from the global model to alleviate the biases between clients. 

\begin{figure}[t]
  \centering
  \includegraphics[width=8cm]{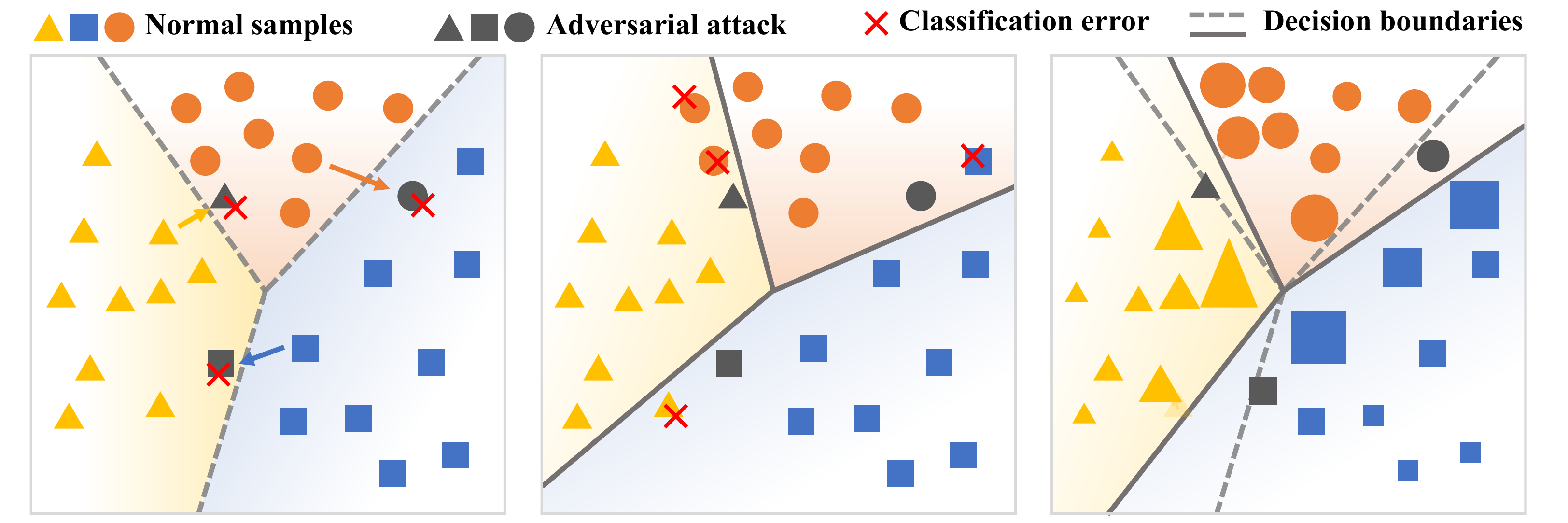}
  
  \caption{\textbf{Left panel:} Decision boundary of plainly trained model. \textbf{Middle panel:} Decision boundary of AT-trained model. \textbf{Right panel:} Decision boundary of DBFAT-trained model. We use the dotted line to represent the boundary of the clean model, and solid line to represent the boundary of the robust model. The size of the shape represents the value of the weight. Those samples that are close to/far from boundary are assigned larger/smaller weight. The decision boundary of DBFAT-trained model (see the right sub-figure) can achieve a higher $A_{rob}$ and meanwhile maintain $A_{cln}$. 
  }
  \label{fig2}
  \vspace{-0.2cm}
\end{figure}

Subsequently, we propose a simple yet effective approach called Decision Boundary based Federated Adversarial Training (DBFAT), which consists of two components. 
For local training, we re-weight adversarial examples to improve robustness; while for global aggregation, we utilize the global model to regularize the accuracy for a lower boundary error $A_{bdy}$. We show the training process of DBFAT in the supplementary and illustrate an example of the 
decision boundary of our approach in Fig.~\ref{fig2}. 





\subsection{Re-weighting with Limited Data}
\label{re-we}
    


 Adversarial examples have the ability to approximately measure the distances from original inputs to a classifier’s decision boundary~\cite{heo2018knowledge}, which can be calculated by the least number of steps that iterative attack (e.g.\ PGD attack~\cite{madry2017towards}) needs in order to find its misclassified adversarial variant.
To better utilize limited adversarial examples, we attempt to re-weight the adversarial examples to guide adversarial training. 
For clean examples that are close to the decision boundary,  we assign larger weights; while those examples that are far from the boundary are assigned with smaller weights.

In this paper, we use PGD-$S$ to approximately measure the geometric distance to the decision boundary, $S$ denotes the number of maximum iteration. We generate adversarial examples as follows~\cite{madry2017towards}:
\begin{equation}
  \begin{aligned}
    {x}^{adv} \leftarrow \Pi_{\mathcal{B}[x, \epsilon]} \left({x}^{adv} +  \alpha \cdot \operatorname{sign}(\nabla_{{x}^{adv}} \ell({x}^{adv}, y)) \right).
  \end{aligned}
\end{equation}

Here $\Pi_{\mathcal{B}[x, \epsilon]}$ is the projection function that
projects the adversarial data back into the $\epsilon$-ball centered at natural data, $\alpha$ is the steps size, $\epsilon$ is perturbation bound.
We find the minimum step $d$, such that after $d$ step of PGD, the adversarial variant can be misclassified by the network, i.e., $arg \; max_c f^{(c)}(x^{adv})\neq y$, where $f^{(c)}(x^{adv})$ is the logits of the $c$-th label.


In this way, given a mini-batch samples $\{(x_i,y_i)\}_{i=1}^m$, then the weight list $\rho$ can be formulated as 
:
\begin{equation}
    \label{eq5}
    \rho \gets 1 - \{ \frac{d_i}{\sum_{i=1}^{m} d_i} \}.
\end{equation}

\subsection{Regularization with Global Model}

Early work~\cite{zhang2019theoretically,cui2021learnable} claims that there exists a trade-off between accuracy and robustness, standard adversarial training can hurt accuracy. To achieve a lower boundary error $A_{bdy}$, we take advantage of logits from the global model $f^{glo}$, which is trained after aggregation. 
Particularly, in federated learning, the model owns the information obtained from the averaged parameters on distributed clients. 

Let $f^{loc}$ denote the adversarially trained model at each local client, $f^{glo}$ has the most desirable classifier boundary for natural data. Then we can modify the  local objective mentioned in Equation \ref{eq1} as below:

\begin{equation}
    \label{eq6}
    \begin{aligned}
      \min \underbrace{ \ell_{ce}(\rho \cdot f^{loc}(x^{adv}),y)}_{\text{for robustness }} +\beta \cdot \underbrace{ \ell_{kl}(f^{loc}(x^{adv}),f^{glo}(x))}_{\text{for accuracy regularization}} .
    \end{aligned}
\end{equation}

Where $\ell_{ce}$ denotes the cross-entropy loss to improve the robustness, and $\ell_{kl}$ is the KL divergence loss to constrain the logits of global model and local model. Here, $\ell_{kl}$ appears as an additional regularization term, which is designed to reduce the boundary error $A_{bdy}=A_{cln}-A_{rob}$. 
Additionally, $\rho$ is the weight calculated by Equation \ref{eq5}, $\beta$ is the parameter to be tuned.

\begin{table}[t]
    \caption{Loss functions of different adversarial training methods.}
    \centering
    \label{loss_adv}
    \scalebox{0.68}{
    \begin{tabular}{cc}
    \toprule
    Defense&Loss Function\\
    \midrule
    PGD\_{AT} & $\operatorname{CE}\left(f\left(x^{adv}\right), y\right)
    $ \\
    ALP & $\operatorname{CE\left(f\left(x^{adv}\right), y\right)}+\beta \cdot\left\|f\left(x^{adv}\right)-f\left(x\right)\right\|_{2}^{2}$ \\
    TRADES & $\operatorname{CE\left(f\left(x\right), y\right)}+\beta \cdot \operatorname{KL}\left(f\left(x^{adv}\right)\|f\left(x\right)\right)$ \\ 
    MMA & $\operatorname{CE\left(f\left(x^{adv}\right), y\right) \cdot \mathbb{R}\left(h_{\boldsymbol{\theta}}(\mathbf{x})=y\right)}+\operatorname{CE\left(f\left(x\right), y\right) \cdot \mathbb{R}\left(h_{\boldsymbol{\theta}}(\mathbf{x})\neq y\right)}$ \\
    AVMixup &  $\operatorname{CE}\left(f\left(x^{av}\right), y^{av}\right) $  \\
    \textbf{DBFAT(ours)} & $\rho \cdot \text{CE}(f(x^{adv}),y) + \beta \cdot \text{KL}\left(f\left(x^{adv}\right)\|f^{glo}\left(x\right)\right)$\\
    \bottomrule
    \end{tabular}}
\end{table}
To show the difference between our DBFAT and existing defense methods, we 
list the loss functions 
of different adversarial training methods in Table \ref{loss_adv}. 

\section{Experimental Results} 
\label{sec5}

\subsection{Experimental Setup} 
\label{setup}
Following the previous work of FL~\cite{mcmahan2017communication}, we distribute training data among 100 clients in both IID and non-IID fashion. For each communication round, we randomly select 10 clients to average the model parameters. All experiments are conducted with 8 Tesla V100 GPUs. More details can be referred to the supplemental material.

\subsubsection{Datasets}
In this section, we show that DBFAT improves the robust generalization and meanwhile maintains a high accuracy with extensive experiments on benchmark CV datasets, including MNIST~\cite{726791}, FashionMNIST~\cite{xiao2017fashionmnist} (FMNIST), CIFAR10~\cite{Krizhevsky_2009_17719}, CIFAR100~\cite{Krizhevsky_2009_17719}, Tiny-ImageNet~\cite{le2015tiny}, and ImageNet-12~\cite{deng2009imagenet}.
The ImageNet-12 is generated  via \cite{li2021anti}, which consists of 12 classes. We resize the original image with size 224*224*3 to 64*64*3 for fast training. 
\subsubsection{Data partitioning} 
\begin{figure}[t]
    \centering
    \includegraphics[width=8.5cm]{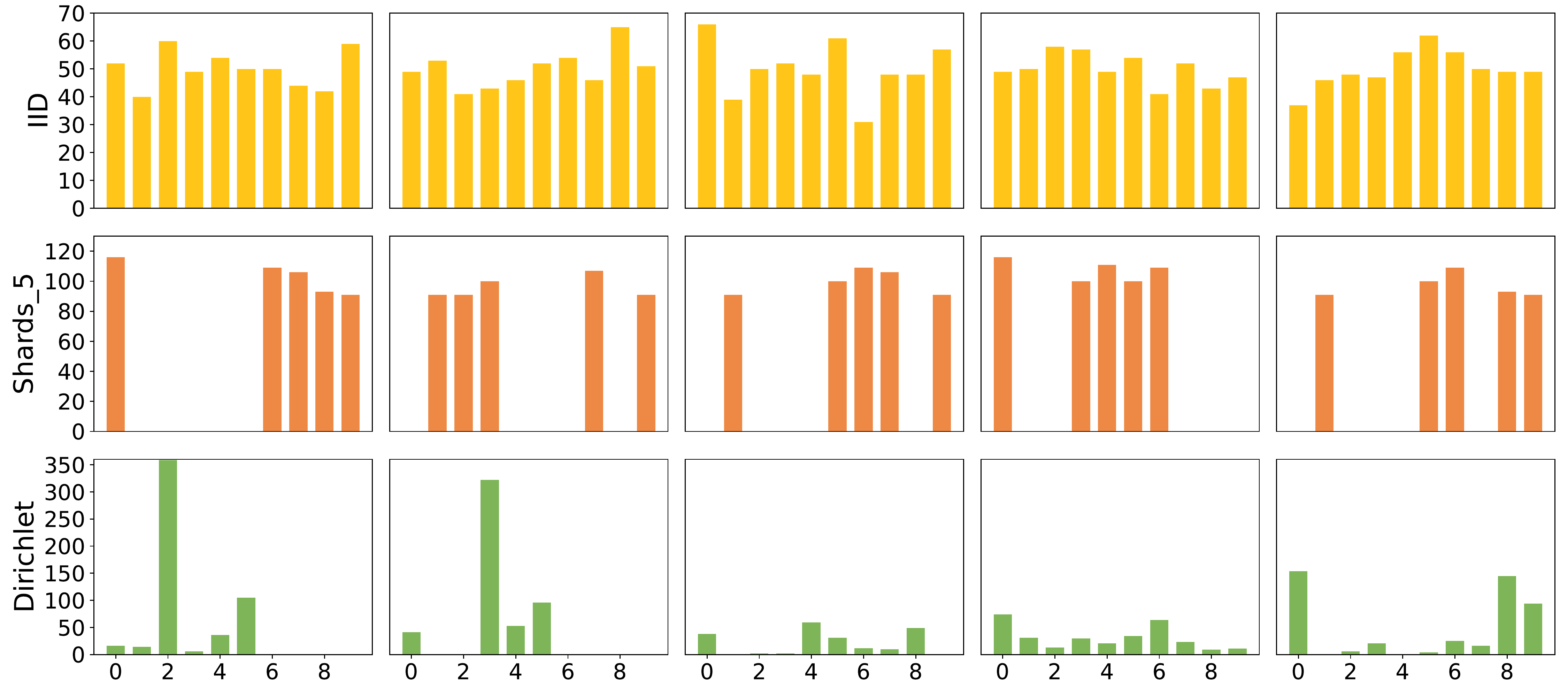}
    \vspace{-0.4cm} 
    \caption{Visualizations of IID and non-IID distribution (Dirichlet sampled and Sharding) across 5 clients on CIFAR10 dataset. 
    Shards\_5 is a type of non-IID setting, in which each client has five categories of data~\cite{mcmahan2017communication}. From left to right: client ID number \#1-5.} 
    \label{illustration}
    \vspace{-0.4cm}
\end{figure}

In the federated learning setup, we evaluate all algorithms on two types of non-IID data partitioning: \textbf{Dirichlet sampled data} and \textbf{Sharding}. For Dirichlet sampled data, each local client is allocated with a proportion of the samples of each label according to Dirichlet distribution~\cite{li2020convergence}. Specifically, we follow the setting in ~\cite{yurochkin2019bayesian}, for each label $c$, we sample $p_c \sim \operatorname{Dir}_{J}(0.5) $ and allocate $p_{c,j}$ proportion
of the whole dataset of label $c$ to client $j$. In this setting, some clients may entirely have no examples of a subset of classes. 
For Sharding~\cite{mcmahan2017communication}, each client owns data samples of a fixed number of labels. Let $K$ be the number of total clients, and $q$ is the number of labels we assign to each client. We divide the dataset by label into $K*q$ shards, and the amount of samples in each shard is $\frac{n}{K\cdot q}$. We denote this distribution as shards\_$q$, where $q$ controls the level of difficulty. If $q$ is set to a smaller value, then the partition is more unbalanced. An example of these partitioning strategies is shown in Fig.~\ref{illustration}, in which we visualize IID and non-IID distribution (Dirichlet sampled with $p_c \sim \operatorname{Dir}_{J}(0.5)$ and Sharding with shards\_$5$) on five randomly selected clients. 
\begin{table*}[t]
\centering
    \caption{Accuracy and adversarial robustness on MNIST, FMNIST and CIFAR10 under both IID and non-IID distribution. An empirical study of FedAvg combined with several defense methods, more detailed comparisons are reported in the supplementary (Section B). Our method significantly outperforms other baselines.}
    \label{jh1}
\scalebox{0.8}{
\begin{tabular}{c|c|cccccc|cccccc}
\toprule
Type &  & \multicolumn{6}{c|}{IID} & \multicolumn{6}{c}{Non-IID} \\
\midrule
Dataset & Method & Clean & FGSM & MIM & PGD-20 & CW & AA & Clean & FGSM & MIM & PGD-20 & CW & AA \\
\midrule
\multirow{6}{*}{MNIST} & Plain & 99.01 & 28.35 & 8.65 & 5.29 & 3.84 & 3.02 & 98.45 & 11.78 & 14.06 & 8.44 & 9.51 & 7.45 \\
 & PGD\_AT & 98.52 & 76.01 & 60.18 & 54.50 & 55.23 & 50.43 & 97.82 & 67.58 & 52.89 & 48.03 & 47.43 & 43.75 \\
 & ALP & 98.46 & 57.37 & 55.61 & 48.74 & 51.17 & 44.25 & 97.92 & 46.49 & 51.01 & 46.41 & 46.24 & 41.95 \\
 & TRADES & 97.89 & 76.79 & 63.29 & 58.25 & 57.24 & 53.72 & 92.03 & 48.45 & 51.56 & 47.21 & 45.81 & 42.36 \\
 & AVMixup & 98.63 & 61.41 & 53.34 & 42.33 & 46.95 & 37.78 & 97.47 & 56.50 & 51.86 & 46.28 & 44.46 & 41.84 \\
 & Ours & \textbf{98.86} & \textbf{78.06} & \textbf{70.97} & \textbf{68.39} & \textbf{63.09} & \textbf{59.39} & \textbf{97.95} & \textbf{68.54} & \textbf{54.18} & \textbf{50.33} & \textbf{49.12} & \textbf{44.32} \\
 \midrule
\multirow{6}{*}{FMNIST} & Plain & 88.50 & 17.89 & 3.55 & 2.57 & 0.40 & 0.17 & 84.60 & 17.86 & 3.25 & 2.93 & 3.05 & -1.40 \\
 & PGD\_AT & 76.05 & 68.53 & 65.24 & 65.40 & 64.26 & 60.89 & 72.93 & 60.11 & 54.42 & 54.33 & 52.19 & 49.88 \\
 & ALP & 75.99 & 67.31 & 63.66 & 63.79 & 61.55 & 59.19 & 75.34 & 57.67 & 53.37 & 55.11 & 51.12 & 51.04 \\
 & TRADES & 78.13 & 59.33 & 52.65 & 52.78 & 51.44 & 48.78 & 74.93 & 56.53 & 44.01 & 44.01 & 31.80 & 39.61 \\
 & AVMixup & 79.34 & 61.22 & 54.93 & 54.67 & 49.48 & 50.07 & 72.06 & 56.26 & 49.21 & 49.72 & 47.99 & 45.15 \\
 & Ours & \textbf{81.49} & \textbf{69.23} & \textbf{66.22} & \textbf{66.24} & \textbf{65.71} & \textbf{61.49} & \textbf{76.19} & \textbf{63.11} & \textbf{56.45} & \textbf{58.31} & \textbf{56.96} & \textbf{53.91} \\
 \midrule
\multirow{6}{*}{CIFAR10} & Plain & 78.80 & 6.87 & 1.15 & 1.06 & 1.30 & 1.23 & 61.10 & 7.58 & 2.94 & 2.67 & 2.87 & 1.28 \\
 & PGD\_AT & 58.75 & 30.62 & 27.23 & 26.11 & 28.47 & 22.09 & 15.27 & 13.27 & 13.00 & 13.00 & 12.99 & 8.63 \\
 & ALP & 63.23 & 29.42 & 26.75 & 28.49 & 28.13 & 23.97 & 32.91 & 21.41 & 20.26 & 20.19 & 17.74 & 15.83 \\
 & TRADES & 68.58 & 31.53 & 25.92 & 25.49 & 23.07 & 20.89 & 46.30 & 24.81 & 22.20 & 22.05 & 19.59 & 17.85 \\
 & AVMixup & 70.28 & 29.51 & 26.22 & 26.34 & 24.07 & 22.25 & 48.23 & 25.29 & 21.42 & 24.25 & 20.25 & 19.43 \\
 & Ours & \textbf{72.21} & \textbf{31.47} & \textbf{28.57} & \textbf{29.03} & \textbf{29.31} & \textbf{24.25} & \textbf{52.24} & \textbf{27.03} & \textbf{24.12} & \textbf{27.02} & \textbf{22.13} & \textbf{21.20} \\
 \bottomrule
\end{tabular}}
\end{table*}

\subsubsection{MNIST and FMNIST setup} We use a simple CNN with two convolutional layers, followed by two fully connected layers. Following the setting used in ~\cite{goodfellow2014explaining}, for MNIST, we set perturbation bound $\epsilon=0.3$, and step size $\alpha=0.01$, and apply adversarial attacks for 20 iterations. For FMNIST, we set perturbation bound $\epsilon=32/255$, and step size $\alpha=0.031$, we adversarially train the network for 10 steps and apply adversarial attacks for 20 iterations. Due to the simplicity of MNIST and FMNIST, we mainly use non-IID data (Sharding), which is hard to train.
\subsubsection{CIFAR10, CIFAR100, Tiny-ImageNet and ImageNet-12 setup} We apply a larger CNN architecture, and follow the setting used in ~\cite{madry2017towards}, i.e., we set the perturbation bound $\epsilon=0.031$, step size $\alpha=0.007$. To evaluate the robustness, we conduct extensive experiments with various data partitioning.
\subsubsection{Baselines} For attack methods, we perform five popular attacks including FGSM~\cite{kurakin2017adversarial}, MIM~\cite{dong2018boosting}, PGD~\cite{madry2017towards}, CW~\cite{carlini2017towards} and AA~\cite{croce2020reliable}. We further use Square~\cite{andriushchenko2020square} for black-box attack. To investigate the effectiveness of existing FL algorithms, we implement FedAvg\cite{mcmahan2017communication}, FedProx\cite{li2018federated}, FedNova\cite{wang2020tackling} and Scaffold\cite{karimireddy2020scaffold}. To defend against adversarial attacks, we implement four most prevailing methods including PGD\_AT\cite{madry2017towards}, TRADES~\cite{zhang2019theoretically}, ALP~\cite{kannan2018adversarial}, MMA~\cite{ding2020mma} and AVMixup~\cite{lee2020adversarial}.  We compare the performance of our DBFAT with various kinds of defense methods combined with FL methods. 




\subsection{Convergence For Local Training} 
\begin{figure}[h]
  \centering
  \begin{minipage}[t]{0.23\textwidth}
    \centering
    \includegraphics[width=4.3cm]{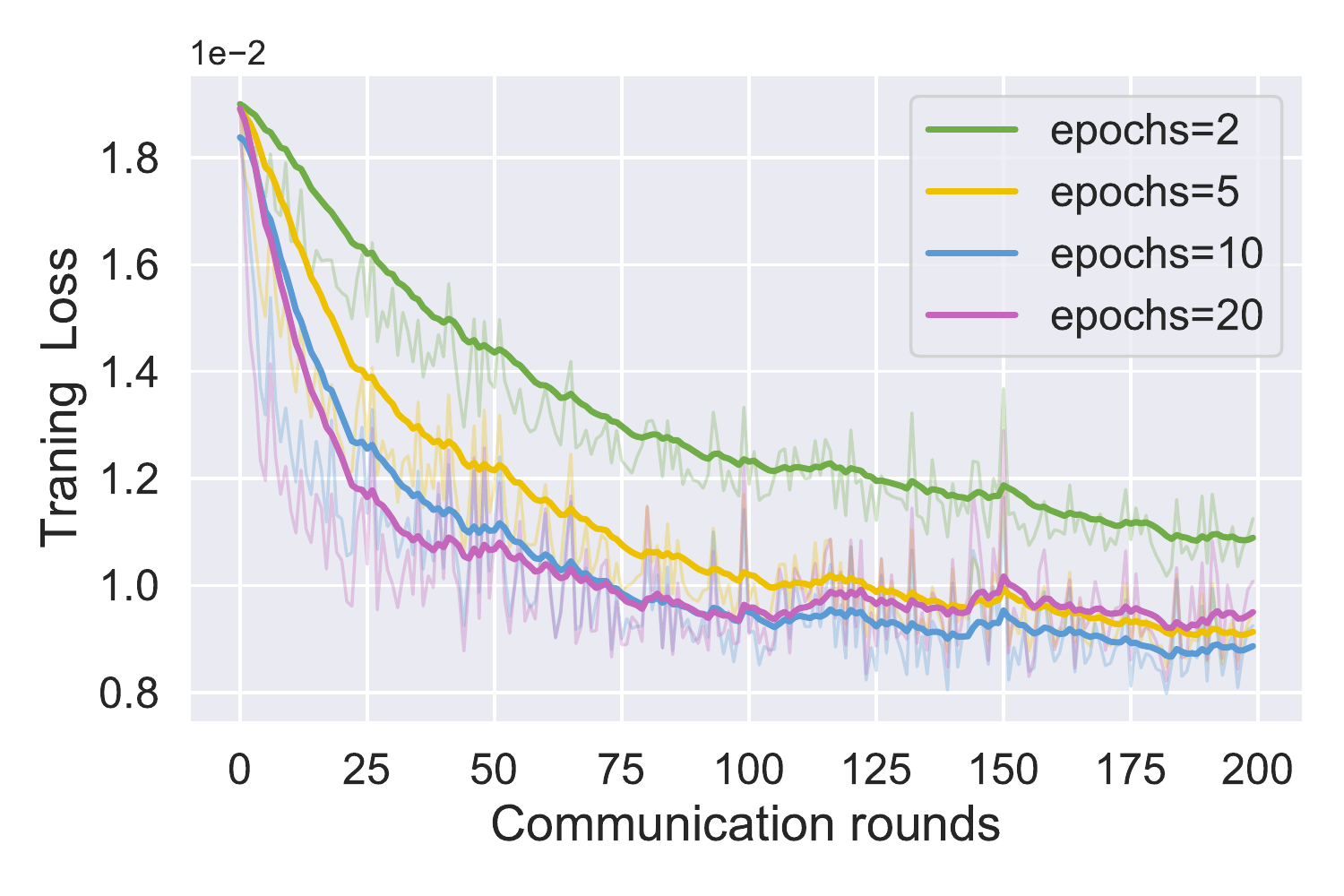}
  \end{minipage} 
  \begin{minipage}[t]{0.23\textwidth}
    \centering
    \includegraphics[width=4cm]{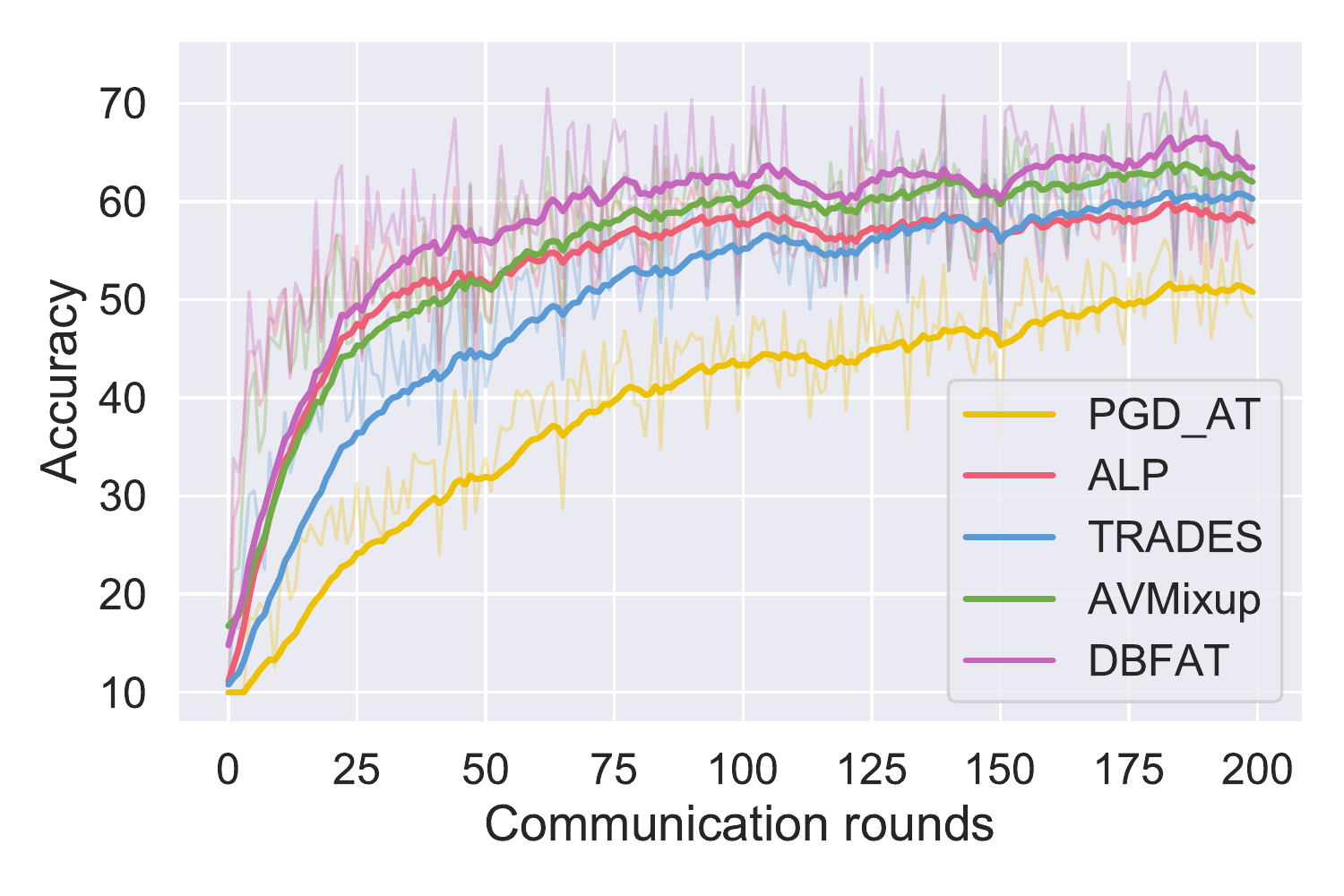}
  \end{minipage}
  \caption{\textbf{Left:} Convergence rate for different local epochs.
  \textbf{Right:} Training curves of FedAvg combined with different AT methods.}
  \label{conv}
\end{figure} 

To show the convergence rate of DBFAT, we use the Dirichlet sampled CIFAR10 dataset, where each client owns 500 samples 
from 5 classes. Fig.~\ref{conv} (left sub-figure) shows the impact of local epoch $E$ during adversarial training. Indeed, for a very small epoch (e.g., $E=2$), it has an extremely slow convergence rate, which may incur more communications. 
Besides, a large epoch (e.g., $E=20$) also leads to a slow convergence, 
as model may overfit to the local data.
Considering both the communication cost and convergence issues, we set $E=5$ in our experiments, which can maintain a proper communication efficiency and fast convergence.
\begin{table*}[t]
\centering
\caption{
Accuracy and adversarial robustness on CIFAR100, Tiny-ImageNet, and ImageNet-12.}
\label{tb:big_data}
\scalebox{0.85}{
\begin{tabular}{c|cccc|cccc|cccc}
\toprule
Dataset & \multicolumn{4}{c|}{CIFAR100} & \multicolumn{4}{c|}{Tiny-ImageNet} & \multicolumn{4}{c}{ImageNet-12} \\
\midrule
Method & Clean & PGD-20 & AA & Square & Clean & PGD-20 & AA & Square & Clean & PGD-20 & AA & Square \\
\midrule
PGD\_AT & 39.32 & 16.07 & 14.36 & 23.44 & 26.33 & 12.26 & 10.26 & 13.54 & 37.42 & 22.61 & 18.30 & 25.57 \\
ALP & 41.12 & 18.46 & 14.78 & 24.54 & 32.78 & 14.62 & 12.19 & 16.48 & 54.96 & 24.78 & 19.57 & 27.73 \\
TRADES & 43.39 & 20.05 & 16.85 & 26.43 & 37.81 & 15.49 & 13.26 & 19.38 & 58.82 & 25.49 & 21.81 & 28.96 \\
AVMixup & 46.64 & 23.56 & 19.46 & 29.16 & 36.19 & 15.28 & 13.18 & 19.25 & 59.63 & 25.81 & 21.92 & 29.28 \\
Ours & \textbf{48.31} & \textbf{24.47} & \textbf{22.46} & \textbf{31.57} & \textbf{38.24} & \textbf{16.17} & \textbf{13.96} & \textbf{20.26} & \textbf{61.38} & \textbf{26.47} & \textbf{22.08} & \textbf{30.91} \\
\bottomrule
\end{tabular}}
\vspace{-0.4cm}
\end{table*}
\subsection{Effectiveness of Our Method}
We verify the effectiveness of our method compared with several adversarial training techniques on Dirichlet sampled CIFAR10. Evaluation of
model robustness is averaged under four attacks using the
the same setting for a fair comparison and all defense methods are combined with FedAvg. 

To show the differences between DBFAT and 
above mentioned defense methods, we report the training curves on non-IID CIFAR10 dataset in the right sub-figure of Fig.~\ref{conv}.   
Fig.~\ref{conv} confirms that our DBFAT achieves the highest clean accuracy.
We speculate that this benefit is due to the regularization term and re-weighting strategy introduced in Equation \ref{eq6}. It is worth mentioning that in the training curves, the model trained with PGD\_AT performs very poorly. It indicates that standard AT may not be a suitable choice for adversarial robustness in FL, as it only uses cross-entropy loss with adversarial examples, but ignores the negative impact on clean accuracy.
 We further report the results on various datasets under both IID and non-IID settings in Table \ref{jh1}, which indicates that DBFAT significantly outperforms other methods in terms of both accuracy and robustness. 
\begin{table}[t]
\centering
\caption{Ablation Study by cutting off different modules.}
\label{jh3}
\scalebox{0.85}{
\begin{tabular}{c|cc|cc}
\toprule
Dataset& \multicolumn{2}{c|}{CIFAR10} & \multicolumn{2}{c}{FMNIST} \\
\midrule
Methods & $A_{cln}$ & Avg $A_{rob}$ & $A_{cln}$ & Avg $A_{rob}$ \\
\midrule
Ours & \textbf{52.16} & \textbf{27.80} & \textbf{75.89} & \textbf{59.63}  \\
\midrule
Ours (w/o re-weighting) & 48.44 & 25.89 & 72.35 &  56.34 \\
\midrule
Ours (w/o regularization) & 51.04 & 26.84  & 73.96 & 58.23\\
\bottomrule
\end{tabular}}
\vspace{-0.3cm}
\end{table}

\subsubsection{Performance on large datasets} 
In Table~\ref{tb:big_data}, we show the accuracy and robustness of each method on large datasets (e.g., CIFAR100, Tiny-ImageNet, and ImageNet-12). All results are tested under PGD-20 attack~\cite{madry2017towards}, AutoAttack~\cite{croce2020reliable}, and Square attack~\cite{andriushchenko2020square} in non-IID settings. From the results reported in Table~\ref{tb:big_data}, we can find that our method still 
outperforms other baselines in terms of both clean accuracy and robustness. Note that our method can achieve the highest accuracy and robustness of 61.38\% and 22.08\% under AutoAttack, respectively. It thus proves that our method can also be used to improve the accuracy and robustness of the model on large datasets.
We think that the higher clean accuracy is a result of the regularization term introduced in Equation~\ref{eq6}, while maintaining a high robustness. 
\subsection{Ablation Study}



\subsubsection{Cutting off different modules}
As part of our ablation study, we first investigate the contributions of different modules introduced in DBFAT. As shown in Table~\ref{jh3}, turning off both the re-weighting strategy and regularization term will lead to poor performance, which demonstrates the
importance of both modules. 
Moreover, cut-offing the re-weighting strategy can lead to a more severe degradation. 
We conjecture this is a reasonable phenomenon. As mentioned in Fig.~\ref{fig1}, non-IID data can cause a serious accuracy reduction. Our re-weighting strategy can alleviate the bias by taking the limited data on each client into account.

\begin{table}[]
\centering
  \caption{Effect of hyper-parameter $\beta$. ``Avg $A_{rob}$" refers to the average robustness under four attacks. }
  \label{jh4}
  \scalebox{0.98}{
  \begin{tabular}{ccccc}
    \toprule
    \multicolumn{1}{l}{Dataset} & \multicolumn{2}{c}{MNIST} & \multicolumn{2}{c}{FMNIST} \\
    \midrule
    $\beta$ & $A_{cln}$ & Avg $A_{rob}$ & $A_{cln}$ & Avg $A_{rob}$ \\
    \midrule
    4 & 98.30\ & 26.64\ & 81.73\ & 37.36\ \\
    2 & 98.14\ & 34.24\ & \textbf{75.59}\ & \textbf{47.83}\ \\
    1.5 & \textbf{98.46}\ & \textbf{53.22}\ & 74.93\ & 44.08\ \\
    1 & 97.32\ & 47.35\ & 65.43\ & 42.33\ \\
    0.5 & 96.57\ & 44.09\ & 61.02\ & 45.28\ \\
    \bottomrule
\end{tabular}}
\vspace{-0.5cm}
\end{table}

\subsubsection{Effects of Regularization}

The regularization parameter $\beta$ is an important hyperparameter in our proposed method. We show how the regularization parameter affects the performance of our robust classifiers by numerical experiments on two datasets, MNIST and FMNIST. In Equation \ref{eq6}, $\beta$ controls the accuracy obtained from the global model, which contains information from distributed clients. Since directly training on adversarial examples could hurt the clean accuracy, here we explore the effects of $\beta$ on both accuracy and robustness. 
As shown in Table \ref{jh4}, we report the clean accuracy and robustness by varying the value of $\beta$.
We empirically choose the best $\beta$ for different datasets. For example, for MNIST, $\beta=1.5$ can achieve better accuracy and robustness. For FMNIST, we let $\beta=2$ for a proper trade-off in accuracy and robustness. 

\section{Conclusion}
\label{cl}

In this paper, we 
investigate an interesting yet not well explored problem in FL: the robustness against adversarial attacks. We first find that directly adopting adversarial training in federated learning can hurt accuracy significantly especially in non-IID setting. We then propose a novel and effective adversarial training method called DBFAT, which is based on the decision boundary of federated learning, and utilizes local re-weighting and global regularization to improve both accuracy and robustness of FL systems. 
Comprehensive experiments on various datasets and detailed comparisons with the state-of-the-art adversarial training methods demonstrate that our proposed DBFAT consistently outperforms other baselines under both IID and non-IID settings.
This work would potentially benefit researchers  who are interested in adversarial robustness of FL. 


\bibliography{aaai23}
\end{document}